# Factored Particles for Scalable Monitoring

Brenda Ng, Leonid Peshkin, Avi Pfeffer*
{bmng,pesha,avi}@eecs.harvard.edu
Division of Engineering and Applied Sciences
Harvard University, Cambridge MA 02138

## Abstract

Exact monitoring in dynamic Bayesian networks is intractable, so approximate algorithms are necessary. This paper presents a new family of approximate monitoring algorithms that combine the best qualities of the particle filtering and Boyen-Koller methods. Our algorithms maintain an approximate representation the belief state in the form of sets of factored particles, that correspond to samples of clusters of state variables. Empirical results show that our algorithms outperform both ordinary particle filtering and the Boyen-Koller algorithm on large systems.

## 1 Introduction

The goal of probabilistic monitoring is to maintain beliefs, in the form of probability distributions over state variables, about the state of a dynamically evolving system. The basic inference step is to produce the new belief state given the previous belief state and the current observations. This step must be performed in real time, and with reasonable accuracy. Monitoring is a difficult problem. Even when a compact representation such as a dynamic Bayesian network is used, the cost of updating the belief state exactly is exponential in the number of state variables. Therefore, approximate monitoring algorithms are needed. This paper presents new algorithms for monitoring in high-dimensional state spaces.

Two main approximate monitoring algorithms have been developed. Both use an approximate representation of the belief state, and propagate that through the dynamics and observations at each time step. In the first algorithm, *particle filtering* [5], the belief state is represented by a set of samples of possible states.

Particle filtering (PF) has the attractive property that the approximation converges to the exact solution in the limit of an infinite number of samples. It also has an "any-real-time" property that the number of samples can be adjusted to fill however much time is available for inference at each time point. However, the variance of the method can be very high, particularly in high dimensional spaces, and too many particles might be needed for decent approximations. The second approximate monitoring algorithm is the *Boyen-Koller* [2] algorithm (BK), which factors the belief state into a product of small terms. Although the factoring introduces error, the expected error does not grow over time. It also performs well in practice for medium sized models. However, a very complex, high-dimensional model may not split reasonably into factors that are small enough to store and propagate in real time.

This paper explores algorithms that combine properties of both the PF and BK algorithms. The basic idea is to represent the belief state using *sets of factored particles*. Each set contains factored particles that specify values for a subset of the state variables. From the point of view of the BK algorithm, a set of factored particles is an approximation to the marginal distribution over the variables in the set. This makes it feasible to store and propagate factors that would otherwise be too large. From the point of view of PF, this method projects the particles down onto subsets of variables at the end of each time step, and then joins these projected particles in the next time step. This *project-join* operation introduces bias into the sampling process, similar to the bias introduced by the BK algorithm. However, this bias is well compensated by decrease in the variance of the filtering process. Our experimental results show that factored particles consistently outperform PF on a variety of models. Furthermore, factored particles attains the benefits of BK on models that are too large for BK itself.

This work was supported by NASA grant NCC2-1236.

## 2　Preliminaries

In this paper, we focus on the task of monitoring discrete-time finite-state Markov processes. We denote the system state at time $t$ as $X_t \in \mathcal{X} = \{x^0, ..., x^n\}$ and the observation as $Y_t \in \mathcal{Y} = \{y^0, ..., y^m\}$. In particular, the states evolve according to a time-invariant Markov process, so that $P(X_t|X_{0:(t-1)}) = P(X_t|X_{(t-1)})$, and $P(Y_t|X_{0:t}) = P(Y_t|X_t)$. With these assumptions, we can characterize the process by its transition model $P(X_t|X_{t-1})$ and its observation model $P(Y_t|X_t)$.

The *belief state* of the system at time $t$ is the probability distribution over the state at time $t$ conditioned upon the sequence of past observations, $P(X_t|Y_{0:t})$. The *monitoring task* is to compute the successive sequence of belief states throughout the lifetime of the system, in real time as the system evolves. Monitoring is therefore a recursive process. The basic step that takes place at time $t$, is to take the previously computed belief state at time $t-1$, incorporate the observation at time $t$, and produce the new belief state at time $t$. Following the notation of [2], we denote the *prior belief state* $\sigma_{\bullet t}$ as the probability distribution over the state at time $t$ conditioned upon the observation sequence up to the last time slice, $\sigma_{\bullet t} \equiv P(X_t|Y_{0:(t-1)})$, and the *posterior belief state* $\sigma_{t \bullet}$ as the probability distribution over the state at time $t$ conditioned upon all the observations including that of the current time slice, $\sigma_{t \bullet} \equiv P(X_t|Y_{0:t})$. Given the posterior belief state $\sigma_{t-1 \bullet}$, the recursive monitoring procedure is as follows:
• Propagate the belief state through the transition model to obtain the *prior belief state* at time $t$:

$$\sigma_{(\bullet t)}(X_t) = \sum_{X_{t-1}} \sigma_{(t-1\bullet)}(X_{t-1}) P(X_t|X_{t-1}) \quad (1)$$

• Condition the prior belief state with the current observation $Y_t$ to obtain the *posterior belief state*:

$$\sigma_{(t\bullet)}(X_t) = \frac{\sigma_{(\bullet t)}(X_t) P(Y_t|X_t)}{\sum_{X_t} \sigma_{(\bullet t)}(X_t) P(Y_t|X_t)}. \quad (2)$$

*Dynamic Bayesian networks* (DBNs) [4] are compact representations of Markov processes. A DBN is a temporal version of a Bayesian network (BN), in which the nodes represent the state of the system at a particular point in time. The set of nodes representing the state at a point in time is called a *time slice*. Nodes in one time slice may have parents in the same time slice and in the previous time slice. A DBN is specified by two components: a prior Bayesian network that represents the initial distribution $\pi_0$ over the initial state, and a *2-time-slice Bayesian network* (2-TBN) that represents the transition distribution from states at time $t$ to states at time $t+1$. A 2-TBN is a fragment of a Bayesian network in which nodes belonging to the previous time slice are roots (i.e. have no parents) and have no conditional probability distributions. It represents the conditional distribution that encodes the transition and observation models of the Markov process.

Although monitoring is a simple procedure in principle, it is quite costly even for processes that are compactly represented as DBNs. In fact, the belief state for a process represented as a DBN is typically exponential in the number of state variables. Therefore, not only is it impossible to represent the belief state exactly as the complete joint probability distribution over the states, it is utterly infeasible to perform exact monitoring.

**The Boyen-Koller algorithm** An approximate algorithm known as the Boyen-Koller (BK) algorithm [2, 3] takes on a parametric approach to approximate the belief state of the system. BK exploits the idea of weak interaction between different system components to artificially impose independencies between weakly-interacting subsystems. BK partitions the state space into subsets or *clusters*, $\mathcal{C} = \{c_1, ..., c_K\}$, of variables, where each $c_i$ corresponds to a subsystem. The belief state over the entire system is represented in factored form, as a set of localized beliefs about the clusters. Formally, the belief state is represented as a product of marginals over the clusters: $P(X_t|Y_{0:t}) \approx \prod_c P(X_t^c|Y_{0:t})$ where $X_t^c$ denotes the set of states under cluster $c$. Thus, the belief state representation in BK is parameterized by the node clustering.

Given an approximate belief state $\hat{\sigma}_{t-1}$, BK derives the next approximate belief state $\hat{\sigma}_t$ by performing one iteration of exact monitoring as outlined above. More precisely, the algorithm computes $\sigma_{\bullet t}$ and $\sigma_{t \bullet}$ using Formulas 1 and 2 above, using $\hat{\sigma}_{t-1}$ as an approximation for $\sigma_{t-1}$. In the DBN framework, this step can be accomplished by applying an exact BN inference algorithm, such as junction tree propagation [9]. In general, the posterior belief state $\sigma_t$ will not fall into the class of factorized distributions that are desired for a decomposed representation. Therefore, BK performs a *projection* step that reduces $\sigma_t$ to the desired factored form $\hat{\sigma}_t$. The projection is accomplished by computing for each cluster $c \in \mathcal{C}$, the marginal under $\sigma_t$ over the variables in $c$.

This projection introduces error into the monitoring process. However, Boyen and Koller have shown that the error due to repeatedly applying this approximation scheme at every time step remains bounded as the process evolves. BK performance is highly sensitive to the clustering. Although small clusters allow for more efficient inference, making clusters too small will increase the error incurred by the factorization. Even if there exists an optimal clustering that reflects the natural decomposition of the physical system, the clusters may contain too many nodes, making them too large to store, or rendering the belief propagation step too costly. Furthermore, even with the simplest possible clustering, in which each cluster consists of a single node, the exact propagation step may prove intractable due to interconnectivity of the model.

The *Factored Frontier* (FF) algorithm [11] is an alternative approximation scheme that attempts to address the intractability of the exact update step. The idea of FF is similar to the fully factored form of BK, which assumes a separate cluster for each node in the network. Instead of doing exact update on the belief state and then project back to the factored representation via marginalization as in BK, FF computes the approximate marginals directly.

**Particle filtering** Particle filtering (PF) [5] is a general purpose Monte Carlo scheme for monitoring dynamic systems by approximating the belief state as a set of samples or *particles*. This nonparametric approach has the advantage that it can approximate any probability distribution and consequently, can be used to monitor systems with changing or uncertain structure. Moreover, PF also has the practical advantage that the amount of work and the resulting accuracy of the monitoring can be tuned finely in accordance with the available computational resources by adjusting the number of particles used to represent the belief state.

Formally, PF is a recursive estimation method that allows for a representation of the posterior belief state $\sigma_{t\bullet} = P(X_t|Y_{0:t})$ as a set of $N$ particles $\{\mathbf{x}_t^{(1)}, ..., \mathbf{x}_t^{(N)}\}$ where each $\mathbf{x}_t^{(i)}$ is a fully instantiated instance of $X_t$. PF approximates the solution to the monitoring problem described above by a discrete sum of weighted particles drawn from the posterior distribution

$$\hat{P}(X_t|Y_{0:t}) \approx \frac{1}{N} \sum_{i=1}^{N} \delta(X_t - \mathbf{x}_t^{(i)}),$$

where $\delta(\cdot)$ denotes the Dirac delta function. The PF process recursively samples a set of particles for each time step, given the previous set of particles and the current observations. Thus the key step is to sample from the conditional distribution $P(X_t|X_{t-1}, Y_t)$. Since it is difficult to sample from this distribution directly, an *importance sampling* method is used. We sample from a more tractable *proposal* distribution and make up the difference by attributing a weight to each particle. The transition distribution $P(X_t|X_{t-1})$ is a standard proposal distribution, and used in our implementation.

Let $X_t^j$ denote the $j^{th}$ node at time t and let $(x_t^j)^{(i)}$ denote the value of $X_t^j$ in the $i^{th}$ particle. The PF algorithm for the DBN framework is as follows [8]:

1. Initialization: $t = 0$
   For $i = 1, ..., N$, traverse the prior BN in topological order and sample each node $X_t^j$ in the network according to the assigned values of its parents.

2. For $t = 1, 2, ...$

   (a) Importance sampling step For $i = 1, ..., N$:
   Initialize $w_t^{(i)} = 1$. Traverse the 2-TBN in topological order. For each node $X_t^j$:
   - If $X_t^j$ is unobserved, sample each node $(x_t^j)^{(i)}$ according to the 2-TBN $P(X_t^j|\mathbf{Pa}[X_t^j]^{(i)})$.
   - If $X_t^j$ is observed, set $(x_t^j)^{(i)}$ to the observed value $x_t^j$. Update the weight $w_t^{(i)} = w_t^{(i)} \cdot P(x_t^j|\mathbf{Pa}[X_t^j]^{(i)})$
   
   *The result of this step is a set of $N$ weighted time-t particles.*

   (b) Resampling step Resample with replacement $N$ particles $\mathbf{x}_t^{(i)}$ with the probability proportional to their weights $w_t^{(i)}$.

An enhancement over PF, known as the *Rao-Blackwellised Particle Filtering* (RBPF) [6], exploits the structure of the DBN to decrease the dimension of the sampling distribution, thus requiring less number of samples to achieve the same degree of accuracy offered by standard PF. The idea of RBPF is that, given a DBN with a *tractable substructure* such that some of nodes can be marginalized out exactly from the posterior distribution, PF can be applied to estimate the distribution over the rest of the nodes which lies in a space of reduced dimension.

## 3 Factored Particles

The main drawback to particle filtering is that the sampling process has high variance, particularly in high-dimensional spaces. As a result, many particles are needed in order to obtain a satisfactory approximation of the belief state. The basic idea behind our method is simple. Instead of maintaining particles over the entire state of the system, we maintain particles over clusters of state variables. Because the clusters have far fewer variables than the entire state space, the variance resulting from maintaining cluster distributions is much smaller than that from maintaining the belief state as a whole. As a result, better approximations can be obtained with smaller numbers of particles. Of course, there is no free lunch—by factoring particles we introduce bias into the belief state approximation. Like the BK method, our method is unable to capture inter-cluster dependencies in the belief state. Our hope is that by introducing a small amount of bias into the sampling process, we can reduce the variance significantly, which would lead to an overall improvement in the accuracy of monitoring. Thus, our method can be viewed as exploiting the *bias-variance* tradeoff [7] that commonly appears in machine learning. The reasons to hope that the bias will be small are the theoretical BK result that the expected error remains bounded in the long run, together with a growing body of experimental evidence showing that the BK error is small. In addition to the published results of Boyen and Koller, we have also found BK to perform very well on a wide range of small models.

Formally, our approximation scheme, called *factored particle filtering* (FP), uses a set of clusters $C = \{c_1, ..., c_K\}$. For each cluster $c$, it maintains a set of *factored particles* $\mathbf{x}_{t,c}^{(1)}, ..., \mathbf{x}_{t,c}^{(N_c)}$, where each factored particle $\mathbf{x}_{t,c}^{(i)}$ is a fully instantiated $X_t^c$, and $N_c$ is the number of particles for $c$.[1] Combining the BK approximation of the belief state as a product of cluster distributions with the PF approximation as a nonparametric density function, we obtain the following approximation of the belief state:

$$\hat{P}(X_t|Y_{0:t}) \approx \prod_C \frac{1}{N_c} \sum_{i=1}^{N_c} \delta\left(X_{t,c} - \mathbf{x}_{t,c}^{(i)}\right).$$

In terms of representation, let the set of nodes in cluster $c$ be $X_1, ..., X_{M_c}$. At each time $t$, each cluster $c$ is associated with a table of factored particles $\{\mathbf{x}_{t,c}\}$, in which each row represents a particular factored particle $\mathbf{x}_{t,c}^{(i)}$.

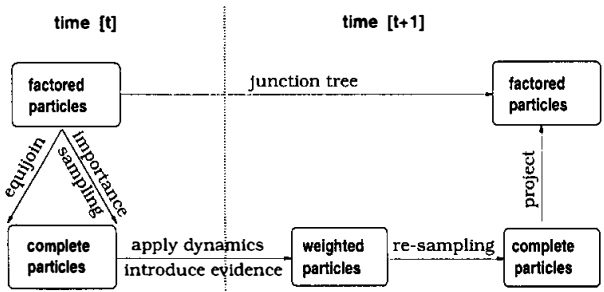

Figure 1: A procedural diagram of the FP algorithms.

The implementation must reproduce a set of such tables at time $t$, given the tables at time $t - 1$ and the observations at time $t$. We provide three implementations of the FP algorithm, that differ in how this is achieved: **FP1**: factored particle filtering within the framework of PF; **FP2**: an efficient approximation to FP1; **FP3**: factored particle filtering using the junction tree. Figure 1 shows a schematic description of the three algorithms. All algorithms begin with sets of factored particles at time $t$, and produce sets of factored particles at time $t+1$. Both FP1 and FP2 take the bottom route, propagating complete particles using the standard PF process. The difference between them is in the way they convert factored particles into full particles. FP1 uses an equijoin operation, while FP2 performs an approximation using importance sampling. Algorithm FP3 takes a direct approach, propagating factored particles through the junction tree and never producing full particles.

---
[1] Although time is not an index to $N_c$, the number of particles for a cluster may vary among time instances.

**Factored Particles within PF** The progress of algorithm FP1 is cyclic. It begins with tables of factored particles, produces complete particles from them, propagates them in the same way as PF to produce new complete particles, and then reduces them back down to factored particles. It is instructive to view the cycle as beginning with complete particles, propagating them through PF to obtain new complete particles, and then performing two operations, going via factored particles to obtain a new set of complete particles. The composition of these two operations is a *project-join* operation, that takes as input a set of $N$ full particles $\mathbf{x}_t^{(i)}$ and returns a new set of $N'$ full particles $\breve{\mathbf{x}}_t^{(i)}$ in which the different clusters are decoupled. In general, $N'$ is larger than $N$. The project-join operator is a composition of two common relational database operators:

**Project** $(\pi_\mathbf{X} R)$ projects the table $R$ onto the attributes $\mathbf{X}$. Let $\mathbf{Y}$ be the attributes other than $\mathbf{X}$ in $R$. Then $\pi_\mathbf{X} R$ has a row $\langle \mathbf{X} = \mathbf{x} \rangle$ for every row $\langle \mathbf{X} = \mathbf{x}, \mathbf{Y} = \mathbf{y} \rangle \in R$. Unlike in standard databases, identical rows are not merged.

**Equijoin** $(R \bowtie S)$ produces a new table over the attributes in $R$ and $S$, in which the attributes appearing in both $R$ and $S$ are forced to be equal. Let $\mathbf{X}$ be the attributes in $R$ and not $S$, $\mathbf{Y}$ the attributes in $S$ and not $R$, and let $\mathbf{Z}$ the attributes in both $R$ and $S$. Then $R \bowtie S$ is a table over $\mathbf{X} \cup \mathbf{Y} \cup \mathbf{Z}$, in which there is a row $\langle \mathbf{X} = \mathbf{x}, \mathbf{Y} = \mathbf{y}, \mathbf{Z} = \mathbf{z} \rangle$ whenever there are rows $\langle \mathbf{X} = \mathbf{x}, \mathbf{Z} = \mathbf{z} \rangle$ in $R$ and $\langle \mathbf{Y} = \mathbf{y}, \mathbf{Z} = \mathbf{z} \rangle \in S$.

For example, if the initial table of complete particles $\langle A, B, C \rangle$ is projected onto the clusters $(A, B)$ and $(B, C)$ as follows:

| A | B | C |
|---|---|---|
| $a_1$ | $b_1$ | $c_1$ |
| $a_2$ | $b_1$ | $c_2$ |
| $a_2$ | $b_2$ | $c_2$ |

$\Rightarrow$

| A | B |
|---|---|
| $a_1$ | $b_1$ |
| $a_2$ | $b_1$ |
| $a_2$ | $b_2$ |

| B | C |
|---|---|
| $b_1$ | $c_1$ |
| $b_1$ | $c_2$ |
| $b_2$ | $c_2$ |

then the equijoin operation produces

| A | B | C |
|---|---|---|
| $a_1$ | $b_1$ | $c_1$ |
| $a_1$ | $b_1$ | $c_2$ |
| $a_2$ | $b_1$ | $c_1$ |
| $a_2$ | $b_1$ | $c_2$ |
| $a_2$ | $b_2$ | $c_2$ |
| $a_2$ | $b_2$ | $c_2$ |

In probabilistic terms, the projection operation marginalizes the distribution represented by the set of full particles to localized distributions over clusters of variables. The equijoin operation is equivalent to taking the product of the marginal distributions, as represented by the sets of factored particles. Thus the project-join operation as a whole is equivalent to taking the product of marginals of the original distribution. This has the effect of breaking all the inter-cluster dependencies, while maintaining the same cluster marginals as the original distribution. It can easily be checked in the above example that $A$ and $C$ are not conditionally independent given $B$ according to the original set of particles, but they are according to the final set. After the projection operation, each of the clusters contains the same number of particles as the original set of full particles. The equijoin operation then produces all the complete particles that are consistent with some factored particle in each cluster. In general, this number will be larger than the original number of particles. However, the total number of particles will not grow from one iteration of the algorithm to the next, because only a certain number of particles will be propagated through the dynamics, and a fixed number will be resampled before the next project-join operation. FP1 is implemented as follows:

1. Initialization: $t = 0$
   Get the first set of full particles as in PF Step 1.

2. For $t = 1, 2, ...$

   - Projection: Let $R_t$ be the table of full particles and let $\mathbf{X}_c$ be the domain of cluster $c$. Project $R_t$ onto each cluster $c$, to produce $R_{t,c} = \pi_{\mathbf{X}_c} R_t$.

   - Equijoin: Join all the factored particle tables together to produce $\breve{R}_t = \bowtie_{c \in C} R_{t,c}$.

   - PF: Perform PF Step 2, beginning with $\breve{R}_t$, to produce the new table of full particles $R_{t+1}$.

**Sampled Equijoin** The feasibility of algorithm FP1 depends on the number of particles produced by the equijoin operation. In general it is considerably more than the original number of particles. The total number depends on the number of clusters and the degree to which clusters are decoupled. In general, with fully decoupled clusters that share no variables, the number of particles produced by equijoin is $N^k$, where $N$ is the original number of particles and $k$ is the number of clusters. This is clearly infeasible with more than a small number of clusters. Our next algorithm is based on the observation that while the equijoin produces the product of marginals of the belief state, there is no reason for us to compute the result exactly. All we are going to do with it is take samples to propagate through the dynamics, to produce the prior distribution at the next instant. Therefore, taking a sample of the product of marginals, and propagating it through the dynamics, should do just as well.

We present an importance sampling method for sampling from the product of marginal distributions. The method starts with the tables of factored particles produced by projection. Then, it processes each of the clusters in turn, gradually creating a new full particle. The particle begins with all variables unset, and weight 1. When a cluster is encountered, some of the cluster variables will already be set. The sample is continued by choosing a row, from the table of factored particles for that cluster, that is consistent with the already chosen variables. To compensate for forcing the sampling process to choose a consistent row, the weight of the sample is multiplied by the fraction of rows in the table for the given cluster that are consistent. Once a row has been chosen for a cluster, all variables in that cluster are set to their value in the row. The FP2 algorithm is the same as FP1, except that the *Equijoin* step is replaced by a *Sample-Join*:

Sample-Join: Let $k$ be the number of clusters, and $N$ the desired number of samples. For $i = 1, 2, \ldots, N$:

- Set $w_t^{(i)} = 1$, and $\mathbf{Z} = \emptyset$

- For $c = 1, 2, ..., k$

  1. Let $\mathbf{X}$ be the variables in cluster $c$, and let $\mathbf{Y} = \mathbf{X} \cap \mathbf{Z}$ Let $R_{t,c}^{\mathbf{y}}$ be the particles in $R_{t,c}$ that agree with the previously chosen values $\mathbf{y}$ for $\mathbf{Y}$. [2]

  2. Uniformly select a factored particle $\mathbf{x}$ from $R_{t,c}^{\mathbf{y}}$ and update : $w_t^{(i)} \leftarrow w_t^{(i)} \times \frac{|R_{t,c}^{\mathbf{y}}|}{|R_{t,c}|}$

  3. $\mathbf{Z} \leftarrow \mathbf{Z} \cup \mathbf{X}$

- Set $\hat{R}_t^{(i)} = \mathbf{z}$, the chosen values for all variables.

---

[2]If $R_{t,c}^{\mathbf{y}}$ is empty, the current sample should be thrown out.

To expedite computation, sample-join is implemented with a preprocessing step that precomputes, for each cluster, the sets of particles consistent with previously selected variables. The preprocessing also prunes away any particles that are never consistent, and computes the weights of the remaining particles. The following example illustrates the sample-join. Given two tables of factored particles:

| A | B | C |
|---|---|---|
| $a_0$ | $b_1$ | $c_0$ |
| $a_0$ | $b_2$ | $c_0$ |
| $a_1$ | $b_1$ | $c_2$ |
| $a_1$ | $b_0$ | $c_3$ |
| $a_2$ | $b_1$ | $c_3$ |
| $a_3$ | $b_2$ | $c_1$ |

| A | C | E |
|---|---|---|
| $a_0$ | $c_0$ | $e_0$ |
| $a_1$ | $c_1$ | $e_0$ |
| $a_2$ | $c_3$ | $e_2$ |
| $a_2$ | $c_3$ | $e_4$ |
| $a_3$ | $c_3$ | $e_1$ |
| $a_3$ | $c_1$ | $e_2$ |

We perform the preprocessing for sample-join:

| A | B | C | w |
|---|---|---|---|
| $a_0$ | $b_1$ | $c_0$ | 1 |
| $a_0$ | $b_2$ | $c_0$ | 1 |
| $a_2$ | $b_1$ | $c_3$ | 1 |
| $a_3$ | $b_2$ | $c_1$ | 1 |

| A | C | E | w |
|---|---|---|---|
| $a_0$ | $c_0$ | $e_0$ | $\frac{1}{4}$ |
| $a_2$ | $c_3$ | $e_2$ | $\frac{1}{2}$ |
| $a_2$ | $c_3$ | $e_4$ | $\frac{1}{2}$ |
| $a_3$ | $c_1$ | $e_2$ | $\frac{1}{4}$ |

The above weights assume that the cluster $ABC$ will be sampled first, and then $ACE$. At the time $ABC$ is sampled, no variables will have assigned values, so any row can be chosen. Therefore the weight is 1. When $ACE$ is sampled, $A$ and $C$ will already have values. Suppose the values $a_0$ and $c_0$ are chosen for $A$ and $C$. Then, in sampling from $ACE$, we will be forced to choose the first row, rather than any of the four rows. To compensate for this, the first row is given weight 1/4. Suppose instead that $a_2$ and $c_3$ were chosen. We would then have a choice of either the second or third row of $ACE$, rather than any of the four rows, so we give them a weight of 1/2 to compensate. The result of taking three samples might be

| A | B | C | E | w |
|---|---|---|---|---|
| $a_0$ | $b_1$ | $c_0$ | $e_0$ | $\frac{1}{8}$ |
| $a_3$ | $b_2$ | $c_1$ | $e_2$ | $\frac{1}{16}$ |
| $a_2$ | $b_1$ | $c_3$ | $e_2$ | $\frac{1}{8}$ |

**Factored Particles in the Junction Tree** Both algorithms FP1 and FP2 use factored particles as stepping stones in the particle filtering process. Most of the time in those algorithms is spent working with full particles. Our third algorithm, FP3, stems from the idea that perhaps the factored particles do not have to be converted into complete particles at all. This is achieved by propagating the tables of factored particles through a junction tree (JT). In BK, a JT is used to propagate the belief state from one instance to the next. This JT is constructed from the 2-TBN. Each cluster of variables in the BK clustering must be contained in two cliques of the JT, one for the previous time slice, and one for the current time slice. We use a JT with the same structure.

In order to take advantage of our sample-based approximation, our JT uses a different representation for potentials from the standard one. In the standard implementation, a potential is a table that assigns a real number to all combinations of values of variables. The size of a potential is exponential in the number of variables. In our implementation, a potential is simply a list of particles, together with weights. A potential can be quite small, even when many variables are involved. Note that ordinary potentials are just special cases of tables of factored particles with weights, in which there is a single particle for each combination of values of the variables. Therefore conditional probability tables can easily be represented with our method. The unweighted particles are simply particles with weights equal to 1.

The JT algorithm requires performing product and summation operations on potentials. In our case, both operations are very simple. The product operation is simply equijoin, where the weight associated with a row is the product of the weights associated with its constituents. The summation operation is simply projection. With these modifications, the JT algorithm is performed in the same way as usual. At each time instance, the following steps are performed:

1. Initialization:
   - Set each of the clique potentials in the JT to the null table over zero variables, which acts as the multiplicative identity.
   - For each conditional probability table in the 2-TBN, choose a clique in the JT containing all of the variables in the table, and multiply the table into the clique potential.
   - For each cluster, choose a clique containing all variables in the cluster at the previous instance, and multiply the table of factored particles into the clique potential.

2. Junction tree calibration, in the standard way.

3. For each cluster, choose a clique in the JT containing all variables in the cluster at the current instance. Marginalize the clique potential onto the cluster variables, to get a table of weighted particles over the cluster variables.

4. For each cluster, resample from the table of weighted particles to get a table of unweighted particles over the cluster variables. This is the new representation of the belief state.

Compared to algorithms FP1 and FP2, we might expect that FP3 would be more accurate. The reason is that FP1 and FP2 involving sampling during the PF propagation phase, whereas in FP3 the propagation is done exactly using the junction tree. The only places sampling is performed in FP3 are at the very first time step, when the first set of factored particles is chosen, and in the resampling step at each iteration.

## 4 Experimental Results

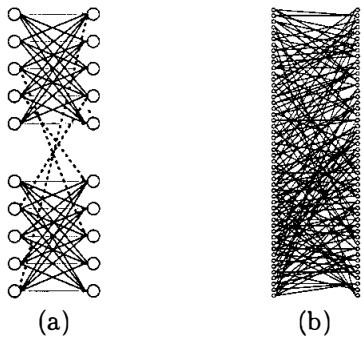

Figure 2: Topologies of experimental networks.

We implemented our algorithms using Kevin Murphy's Matlab toolbox [10]. Our experiments are preliminary, but the results are encouraging. In our first set of experiments, we compared our algorithms to PF and BK on a number of fairly small network topologies. Our family of FP algorithms consistently outperformed the regular PF, but both were dominated by BK on small networks with the fairly small number of particles that we used. A typical topology, shown in Figure 2(a) contained two strongly connected clusters of five nodes, with a small number random connections between clusters. All variables are binary, and conditional probability distributions are random.

The results are shown in Figure 3, which presents the KL-distance between approximate belief states obtained by several algorithms, and the exact belief state. The KL-distance is shows as a function of $t$, the iteration of the algorithm. The results are averaged over 50 trials. As can be seen in the graph, there is a "burn-in" period, followed by stability in the average error. All three of the FP algorithms ourperformed PF, and FP3 was the best of the FP algorithms. All the algorithms were given approximately the same running time per iteration. The performance of BK is not shown in the graph. BK significantly outperformed the other algorithms in accuracy, and also took less time than they were given. Qualitatively similar results are obtained when one compares the probability assigned by the exact and approximate algorithms to the observed values.

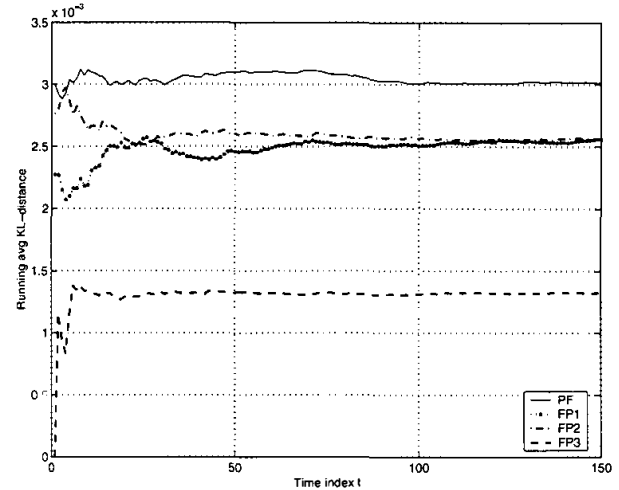

Figure 3: Comparison of the KL-distance between approximate and exact distributions.

All this experiment shows is that when the model strongly supports clustering, exploiting it can lead to an improvement over the performance of PF. It is not surprising that BK would perform extremely well in such a system. However, our hope is that factored particles can be beneficial in much larger systems where BK is infeasible, and even in systems that do not have an obvious clustering. To test this, we tried our algorithms on DBNs such as that shown in Figure 2(b). This model has 50 nodes, and each node has three random parents in the previous time slice. There is no clustering. The nodes are again binary, but the conditional probability tables are skewed so that entries are close to 0 or 1.

| method | # of clusters | $-\log lik.$ | time/iter. |
|--------|---------------|--------------|------------|
| PF     | -             | 145.36       | 3.1 sec.   |
| FP1    | 2             | 120.45       | 2.7 sec.   |
| FP2    | 2             | 138.16       | 2.5 sec.   |
| FP2    | 3             | 127.23       | 2.6 sec.   |
| FP2    | 4             | **103.45**   | 2.7 sec.   |
| FP2    | 6             | 123.64       | 2.8 sec.   |
| FP2    | 12            | 132.27       | 3.2 sec.   |

Table 1: Performance of FP2 on various clusterings in a large DBN.

Because the model is so large, we were unable to run exact inference to obtain ground truth. Instead, we measured, for each algorithm, the negative log likelihood of the observations, according to the belief state produced the algorithm. This is a surrogate for the error of the algorithm, since an inaccurate algorithm will be more surprised by the observations, and therefore have a higher negative log likelihood. Also, BK could not be performed on this model, even when the clus-ters consist of individual nodes. The 2-TBN is highly multi-connected, so the junction tree has very large clusters that cannot be allocated. Because our implementation of FP3 built on top of a standard junction tree implementation that allocated memory for all the cliques, we also could not run it on this model. However, with a special-purpose junction tree implementation in which memory was only allocated to store the tables as needed, we believe FP3 could work on this model.

The results for PF, FP1 and FP2 are shown in Table 1. In addition to the negative log likelihood, the table shows the running time per iteration that each algorithm took. FP1 outperformed FP2 when there were only two clusters, which is natural since FP2 is an approximation to FP1. However, FP1 could not be run with more clusters, due to the large number of equijoin rows produced. We see that in fact there is an optimal number of clusters, and FP2 with 4 clusters significantly outperforms all other algorithms on this model. It is not surprising that there is an optimal number of clusters — it represents the optimal point on the bias-variance tradeoff.

## 5 Discussion and Future Work

There are many open theoretical questions about the FP algorithms. Both PF and BK have theoretical guarantees as to the quality of the approximation, but they are of a very different nature. Generic PF has the nice property that in the limit of an infinite number of particles, the approximate representation of the belief state approaches the exact belief state (see theorems 4.3.1 and 4.3.2 in [1]). One would guess that in the limit of an infinite number of particles, the approximate representation produced by each of the FP algorithms converges to the representation of the BK algorithm. If so, one could then invoke the result that BK has limited expected error to provide a similar guarantee for FP with a large number of samples.

Further questions concern what happens with a finite number of samples. Is it the case that the error introduced by the projection step using our method is bounded over time, in the same way that the BK error is bounded over time? More concretely, can we characterize the bias-variance tradeoff between PF and FP in a formal way?

One of the attractive features of PF is that it can be applied to continuous and hybrid systems. We would like to extend the FP algorithms so that they can handle continuous domains. The main issue is defining the join operation on tables of factored particles: when variables have continuous domains, there will likely be no values in common between the two tables.

The FP algorithms inherit the any-real-time nature of PF: the number of particles used by each of the algorithms can be adjusted to the amount of computation time available on each iteration. Furthermore, our algorithm has an additional attractive property that the computation time can be budgeted between different subsystems, by allocating different numbers of particles to the different systems. We would like to investigate architectures for performing this allocation of resources. This would require algorithms to determine how many particles to allocate to a subsystem based on factors such as the importance of the subsystem and the volatility of its dynamics. Ideally, we would like to distribute resources on the fly, in response to perceived events.

## References

[1] C. Andrieu, A. Doucet, and E. Punskaya. Sequential monte carlo methods for optimal ltering. In J. F. G. d. F. A. Doucet and N. J. Gordon, editors, *Sequential Monte Carlo Methods in Practice*. Springer-Verlag, 2000.

[2] X. Boyen and D. Koller. Tractable inference for complex stochastic processes. In *Proceedings of the Fourteenth Conference on Uncertainty in Artificial Intelligence*, pages 33–42, 1998.

[3] X. Boyen and D. Koller. Exploiting the architecture of dynamic systems. In *Proceedings of the Sixteenth National Conference on Artificial Intelligence*, pages 313–320, 1999.

[4] T. Dean and K. Kanazawa. A model for reasoning about persistence and causation. *Computational Intelligence*, 5:142–150, 1989.

[5] A. Doucet. On sequential simulation-based methods for bayesian filtering. Technical Report CUED/F-INFENG/TR. 310, Cambridge University Department of Engineering, 1998.

[6] A. Doucet, N. de Freitas, K. Murphy, and S. Russell. Rao-blackwellised particle filtering for dynamic Bayesian networks. In *Proceedings of the 16th Conference on Uncertainty in Artificial Intelligence*, 2000.

[7] S. Geman, E. Bienenstock, and R. Doursat. Neural networks and the bias/variance dilemma. *Neural Computation*, 4:1–58, 1992.

[8] D. Koller and U. Lerner. *Sequential Monte Carlo Methods in Practice*, chapter Sampling in Factored Dynamic Systems, pages 445–464. Springer, 2001.

[9] S. L. Lauritzen and D. J. Spiegelhalter. Local computations with probabilities on graphical structures and their application to expert systems. *Journal of the Royal Statistical Society*, pages 157–224, 1988.

[10] K. Murphy. The Bayes net toolbox for Matlab. *Computing Science and Statistics*, 2001.

[11] K. Murphy and Y. Weiss. The factored frontier algorithm for approximate inference in DBNs. In *NIPS*, volume 12, 2000.